\pdfoutput=1
\documentclass[11pt]{article}

\usepackage[]{packages/acl}

\usepackage{times}
\usepackage{latexsym}
\usepackage[T1]{fontenc} 
\usepackage[utf8]{inputenc}
\usepackage{microtype} 

\usepackage[pdftex]{graphicx}           
\usepackage{subcaption}                 
\usepackage{amsmath}                    
\usepackage{amsfonts}                   
\usepackage{cleveref}                   
\usepackage{multirow}                   
\usepackage{pifont}                     
\usepackage[show]{packages/chato-notes} 
\usepackage{listings}                   
\usepackage{scalerel,xparse}            

\newcommand{\at}{\texttt{@}}
\newcommand{\cmark}{\ding{51}}%
\newcommand{\xmark}{\ding{55}}%
\newcommand{\hf}{\includegraphics[scale=0.08]{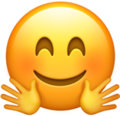} }
\lstdefinestyle{CEE}{language=Python, frame=l,  numbers=left,  numbersep=1em,  xleftmargin=2em, basicstyle=\footnotesize}

\title{A Statutory Article Retrieval Dataset in French}

\author{%
    Antoine Louis \and Gerasimos Spanakis \\
    Law \& Tech Lab, Maastricht University \\
    {\small \texttt{\{a.louis, jerry.spanakis\}@maastrichtuniversity.nl}} \\
}

\begin{document}
\maketitle

\begin{abstract}
  Statutory article retrieval is the task of automatically retrieving law articles relevant to a legal question. While recent advances in natural language processing have sparked considerable interest in many legal tasks, statutory article retrieval remains primarily untouched due to the scarcity of large-scale and high-quality annotated datasets. To address this bottleneck, we introduce the Belgian Statutory Article Retrieval Dataset (BSARD), which consists of 1,100+ French native legal questions labeled by experienced jurists with relevant articles from a corpus of 22,600+ Belgian law articles. Using BSARD, we benchmark several state-of-the-art retrieval approaches, including lexical and dense architectures, both in zero-shot and supervised setups. We find that fine-tuned dense retrieval models significantly outperform other systems. Our best performing baseline achieves 74.8\% R\at100, which is promising for the feasibility of the task and indicates there is still room for improvement. By the specificity of the domain and addressed task, BSARD presents a unique challenge problem for future research on legal information retrieval. Our dataset and source code are publicly available.
\end{abstract}

\section{Introduction \label{sec:introduction}}
Legal issues are an integral part of many people’s lives \citep{long2019global}. However, the majority of citizens have little to no knowledge about their rights and fundamental legal processes \citep{balmer2010knowledge}. As the Internet has become the primary source of information in response to life problems \citep{estabrook2007information}, people increasingly turn to search engines when faced with a legal issue \citep{denvir2016online}. Nevertheless, the quality of the search engine’s legal help results is currently unsatisfactory, as top results mainly refer people to commercial websites that provide basic information as a way to advertise for-profit services \citep{hagan2020legal}. On average, only one in five persons obtain help from the Internet to clarify or solve their legal issue \citep{long2019global}. As a result, many vulnerable citizens who cannot afford a legal expert’s costly assistance are left unprotected or even exploited. This barrier to accessing legal information creates a clear imbalance within the legal system, preventing the right to equal access to justice for all. 

People do not need legal services in and of themselves; they need the ends that legal services can provide. Recent advances in natural language processing (NLP), combined with the increasing amount of digitized textual data in the legal domain, offer new possibilities to bridge the gap between people and the law. For example, legal judgment prediction \citep{aletras2016predicting,luo2017learning,zhong2018legal,hu2018few,chen2019charge} may assist citizens in finding insightful patterns between their case and its outcome. Additionally, legal text summarization \citep{hachey2006extractive,bhattacharya2019comparative} and automated contract review \citep{harkous2018polisis,lippi2019claudette} may help people clarify long, complex, and ambiguous legal documents. 

\begin{figure*}[t]
    \centering
    \includegraphics[width=1\linewidth]{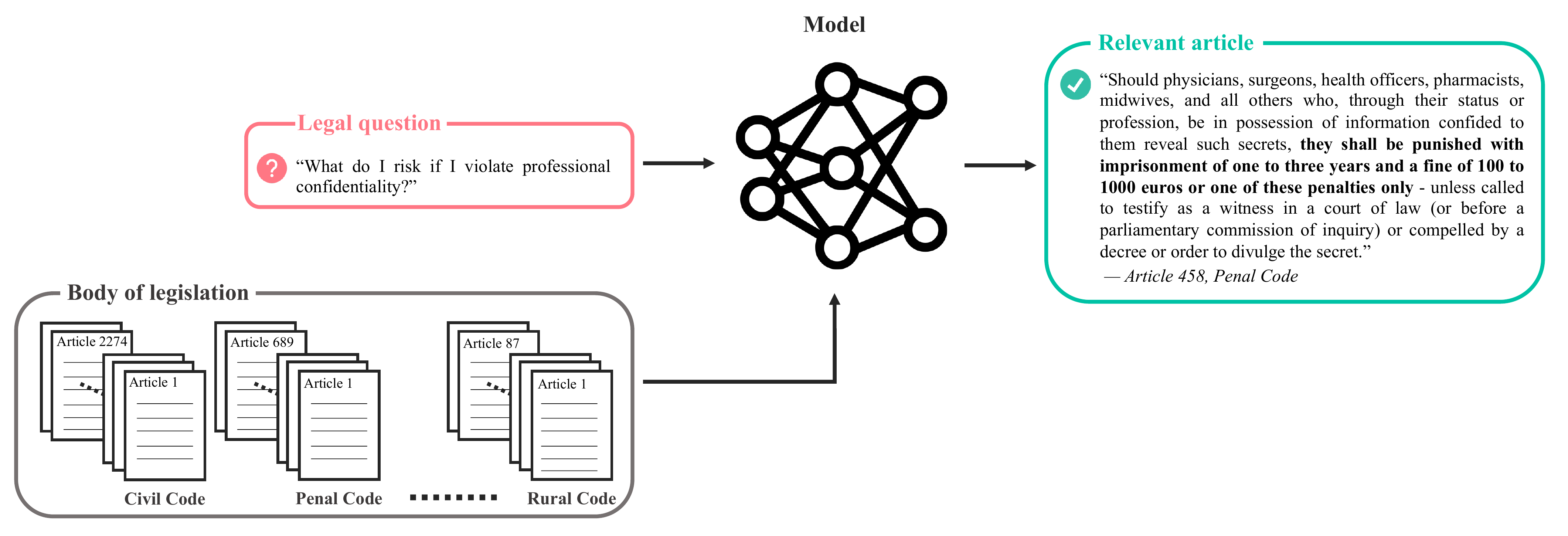}
    \caption{Illustration of the statutory article retrieval task performed on the Belgian Statutory Article Retrieval Dataset (BSARD), which consists of 1,100+ questions carefully labeled by legal experts with references to relevant articles from the Belgian legislation. With BSARD, models can learn to retrieve law articles relevant to a legal question. All examples we show in the paper are translated from French for illustration.}
    \label{fig:task}
\end{figure*}

In this work, we focus on statutory article retrieval, which, given a legal question such as “\textsl{Is it legal to contract a lifetime lease?}”, aims to return one or several relevant law articles from a body of legal statutes \citep{kim2019statute,nguyen2020jnlp}, as illustrated in \Cref{fig:task}. A qualified statutory article retrieval system could provide a professional assisting service for unskilled humans and help empower the weaker parties when used for the public interest.

Finding relevant statutes to a legal question is a challenging task. Unlike traditional ad-hoc information retrieval \citep{craswell2020overview}, statutory article retrieval deals with two types of language: common \textit{natural} language for the questions and complex \textit{legal} language for the statutes. This difference in language distribution greatly complicates the retrieval task as it indirectly requires an inherent interpretation system that can translate a natural question from a non-expert to a legal question to be matched against statutes. For skilled legal experts, these interpretations come from their knowledge of a question’s domain and their understanding of the legal concepts and processes involved. Nevertheless, an interpretation is rarely unique. Instead, it is the interpreter’s subjective belief that gives meaning to the question and, accordingly, an idea of the domains in which the answer can be found. As a result, the same question can yield different paths to the desired outcome depending on its interpretation, making statutory article retrieval a difficult and time-consuming task. 

Besides, statutory law is not a stack of independent articles to be treated as complete sources of information on their own -- unlike news or recipes. Instead, it is a structured and hierarchical collection of legal provisions that have whole meaning only when considered in their overall context, i.e., together with the supplementary information from their neighboring articles, the fields and sub-fields they belong to, and their place in the hierarchy of the law. For instance, the answer to the question “\textsl{Can I terminate an employment contract?}” will most often be found in labor law. However, this is not necessarily true if an employer is contracting a self-employed worker to carry out a specific task, in which case the answer probably lies at the higher level of contract law. This example illustrates the importance of considering the question’s context and understanding the hierarchical structure of the law when looking for relevant statutory articles.

In order to study whether retrieval models can approximate the efficiency and reliability of legal experts, we need a suitable labeled dataset. However, such datasets are difficult to obtain considering that, although statutory provisions are generally publicly accessible (yet often not in a machine-readable format), the questions posed by citizens are not. 

This work presents a novel French native expert-annotated statutory article retrieval dataset as its main contribution. Our Belgian Statutory Article Retrieval Dataset (BSARD) consists of more than 1,100 legal questions posed by Belgian citizens and labeled by legal experts with references to relevant articles from a corpus of around 22,600 Belgian law articles. As a second contribution, we establish strong baselines on BSARD by comparing diverse state-of-the-art retrieval approaches from lexical and dense architectures. Our results show that fine-tuned dense retrieval models significantly outperform other approaches yet suggest ample opportunity for improvement. We publicly release our dataset and source code at {\footnotesize \url{https://github.com/maastrichtlawtech/bsard}}.

\section{Related Work \label{sec:related-work}}
Due to the increasing digitization of textual legal data, the NLP community has recently introduced more and more datasets to help researchers build reliable models on several legal tasks. For instance, \citet{fawei2016passing} introduced a legal question answering (LQA) dataset with 400 multi-choices questions based on the US national bar exam. Similarly, \citet{zhong2020jec} released an LQA dataset based on the Chinese bar exam consisting of 26,365 multiple-choice questions, together with a database of evidence that includes 3,382 Chinese legal provisions and the content of the national examination counseling book. 

Furthermore, \citet{duan2019cjrc} proposed a legal reading comprehension dataset with 52,000 question-answer pairs crafted on the fact descriptions of 10,000 cases from the Supreme People’s Court of China. On a different note, \citet{xiao2018large} presented a dataset for legal judgment prediction (LJP) with around 2.68 million Chinese criminal cases annotated with 183 law articles and 202 charges. Likewise, \citet{chalkidis2019neural} introduced an LJP dataset consisting of 11,478 English cases from the European Court of Human Rights labeled with the associated final decision. 

Meanwhile, \citet{xiao2019dataset} introduced a dataset for similar case matching with 8,964 triplets of cases published by the Supreme People’s Court of China, and \citet{chalkidis2019large} released a text classification dataset containing 57,000 English EU legislative documents tagged with 4,271 labels from the European Vocabulary. Additionally, \citet{manor2019plain} introduced a legal text summarization dataset consisting of 446 sets of contract sections and corresponding reference  summaries, and \citet{holzenberger2020dataset} presented a statutory reasoning dataset based on US tax law. 

Recently, \citet{hendrycks2021cuad} proposed a dataset for legal contract review that includes 510 contracts annotated with 41 different clauses for a total of 13,101 annotations. In the same vein, \citet{borchmann2020contract} introduced a semantic retrieval dataset for contract discovery with more than 2,500 annotations in around 600 documents. Lastly, the COLIEE Case Law Corpus \citep{rabelo2020coliee} is a case law retrieval and entailment dataset that includes 650 base cases from the Federal Court of Canada, each with 200 candidate cases to be identified as relevant to the base case.

Regarding statutory article retrieval, the only other publicly available dataset is the COLIEE Statute Law Corpus \citep{rabelo2020coliee}. It comprises 696 questions from the Japanese legal bar exam labeled with references to relevant articles from the Japanese Civil Code, where both the questions and articles have been translated from Japanese to English. However, this dataset focuses on legal bar exam question answering, which is quite different from legal questions posed by ordinary citizens. While the latter tend to be vague and straightforward, bar exam questions are meant for aspiring lawyers and are thus specific and advanced. Besides, the dataset only contains closed questions (i.e., questions with “yes” or “no” answers) and considers almost 30 times fewer law articles than BSARD does. Also, unlike BSARD, the data are not native sentences but instead translated from a foreign language with a completely different legal system.\footnote{Japan is a civil law country that relies predominantly on the rules written down in statutes, whereas most English-speaking countries (e.g., US, UK, Canada, and Australia) have a common law system that relies predominantly on past judicial decisions, known as precedents.} As a result, the translated dataset may not accurately reflect the logic of the original legal system and language. These limitations suggest the need for a novel large-scale citizen-centric native dataset for statutory article retrieval, which is the core contribution of the present work.

\section{The Belgian Statutory Article Retrieval Dataset \label{sec:dataset}}

\subsection{Dataset Collection \label{subsec:dataset-collection}}
We create our dataset in four stages: (i) compiling a large corpus of Belgian law articles, (ii) gathering legal questions with references to relevant law articles, (iii) refining these questions, and (iv) matching the references to the corresponding articles from our corpus.

\paragraph{Law articles collection.}
In civil law jurisdictions, a legal code is a type of legislation that purports to exhaustively cover a whole area of law, such as criminal law or tax law, by gathering and restating all the written laws in that area into a unique book. Hence, these books constitute valuable resources to collect many law articles on various subjects. We consider 32 publicly available Belgian codes, as presented in \Cref{tab:codes} of \Cref{app:legal_codes}. Together with the legal articles, we extract the corresponding headings of the sections in which these articles appear (i.e., book, part, act, chapter, section, and subsection names). These headings provide an overview of each article’s subject. As pre-processing, we use regular expressions to clean up the articles of specific wording indicating a change in part of the article by a past law (e.g., nested brackets, superscripts, or footnotes). Additionally, we identify and remove the articles repealed by past laws but still present in the codes. Eventually, we end up with a corpus $\mathcal{C} = \{a_1, \cdots, a_N\}$ of $N=22,633$ articles that we use as our basic retrieval units.

\paragraph{Questions collection.}
We partner with Droits Quotidiens (DQ),\footnote{\url{https://droitsquotidiens.be/}} a Belgian organization whose mission is to clarify the law for laypeople. Each year, DQ receives and collects around 4,000 emails from Belgian citizens asking for advice on a personal legal issue. Thanks to these emails, its team of six experienced jurists keeps abreast of Belgium’s most common legal issues and addresses them as comprehensively as possible on its website. Each jurist is an expert in a specific field (e.g., “family”, “housing”, or “work”) and is responsible for answering all questions related to that field. Given their qualifications and years of experience in providing legal advice in their respective fields, the experts can be considered competent enough to always (eventually) retrieve the correct articles to a given question. 

In practice, their legal clarification process consists of four steps. First, they identify the most frequently asked questions on a common legal issue. Then, they define a new anonymized “model” question on that issue expressed in natural language terms, i.e., as close as possible as if a layperson had asked it. Next, they search the Belgian law for articles that help answer the model question and reference them. Finally, they answer the question using the retrieved relevant articles in a way a layperson can understand. These model questions, legal references, and answers are further categorized before being posted on DQ’s website (e.g., the question “\textsl{What is the seizure of goods?}” is tagged under the “\textsl{Money $\rightarrow$ Debt recovery}” category). With their consent, we collect more than 3,200 model questions together with their references to relevant law articles and categorization tags. 

Assuming it takes a jurist between 5 to 20 minutes to find the relevant articles to a given question and categorize the latter. An estimate of the pecuniary value of those labeled questions is over €105,000 -- 3,200 questions, each requiring 10 minutes to label, assuming a rate of €200 per hour.

\paragraph{Questions refinement.}
We find that around one-third of the collected questions are duplicates. However, these duplicated questions come with different categorization tags, some of which providing additional context that can be used to refine the questions. For example, the question ``\textsl{Should I install fire detectors?}'' appears four times in total, under the following tags: ``\textsl{Housing $\rightarrow$ Rent $\rightarrow$ I am a \{\textit{tenant}, \textit{landlord}\} $\rightarrow$ In \{\textit{Wallonia}, \textit{Brussels}\}}''. We distinguish between the tags with one or a few words indicating a question \textit{subject} (e.g., ``\textsl{housing}'' and ``\textsl{rent}'') and those that provide \textit{context} about a personal situation or location as short descriptive sentences (e.g., ``\textsl{I am tenant in Brussels.}''). If any, we append the contextual sentence tags in front of the questions, which solves most of the duplicates problem and improves the overall quality of the questions by making them more specific.

\paragraph{Questions filtering.}
The questions collected are annotated with plain text references to relevant law articles (e.g., “\textsl{Article 8 of the Civil Code}”). We use regular expressions to parse these references and match them to the corresponding articles from our corpus. First, we filter out questions whose references are not articles (e.g., an entire decree or order). Then, we remove questions with references to legal acts other than codes of law (e.g., decrees, directives, or ordinances). Next, we ignore questions with references to codes other than those we initially considered. We eventually end up with 1,108 questions, each carefully labeled with the ids of the corresponding relevant law articles from our corpus. Finally, we split the dataset into training/test sets with 886 and 222 questions, respectively.

\subsection{Dataset Analysis \label{subsec:dataset-analysis}}
To provide more insight, we describe quantitative and qualitative observations about BSARD. Specifically, we explore (i) the diversity in questions and articles, (ii) the relationship between questions and their relevant articles, and (iii) the type of reasoning required to retrieve relevant articles.

\begin{table*}[t]
\centering
\resizebox{\textwidth}{!}{%
\begin{tabular}{lrll}
\hline
\textbf{General topic}  & \textbf{Percentage}   & \textbf{Subtopics}                     & \textbf{Example}         \\ 
\hline
Family                  & 30.6\%                & Marriage, parentage, divorce, etc.     & \textsl{When is there a guardianship?}         \\
Housing                 & 27.4\%                & Rental, flatshare, insalubrity, etc.   & \textsl{Who should repair the common wall?}     \\
Money                   & 16.0\%                & Debts, insurance, taxes, etc.          & \textsl{What is the seizure of goods?}         \\
Justice                 & 13.6\%                & Proceedings, crimes, legal aid, etc.   & \textsl{How does the appeal process work?}     \\
Foreigners              & 5.7\%                 & Citizenship, illegal stay, etc.        & \textsl{Can I come to Belgium to get married?} \\
Social security         & 3.5\%                 & Pensions, pregnancy, health, etc.      & \textsl{Am I dismissed during my pregnancy?}   \\
Work                    & 3.2\%                 & Breach of contract, injuries, etc.     & \textsl{Can I miss work to visit the doctor?}  \\
\hline
\end{tabular}%
}
\caption{Distribution of question topics in BSARD.}
\label{tab:question-topics}
\end{table*}

\begin{figure*}[t]
\centering
\resizebox{\textwidth}{!}{%
\begin{subfigure}[t]{.22\linewidth}
    \centering
    \includegraphics[width=\linewidth]{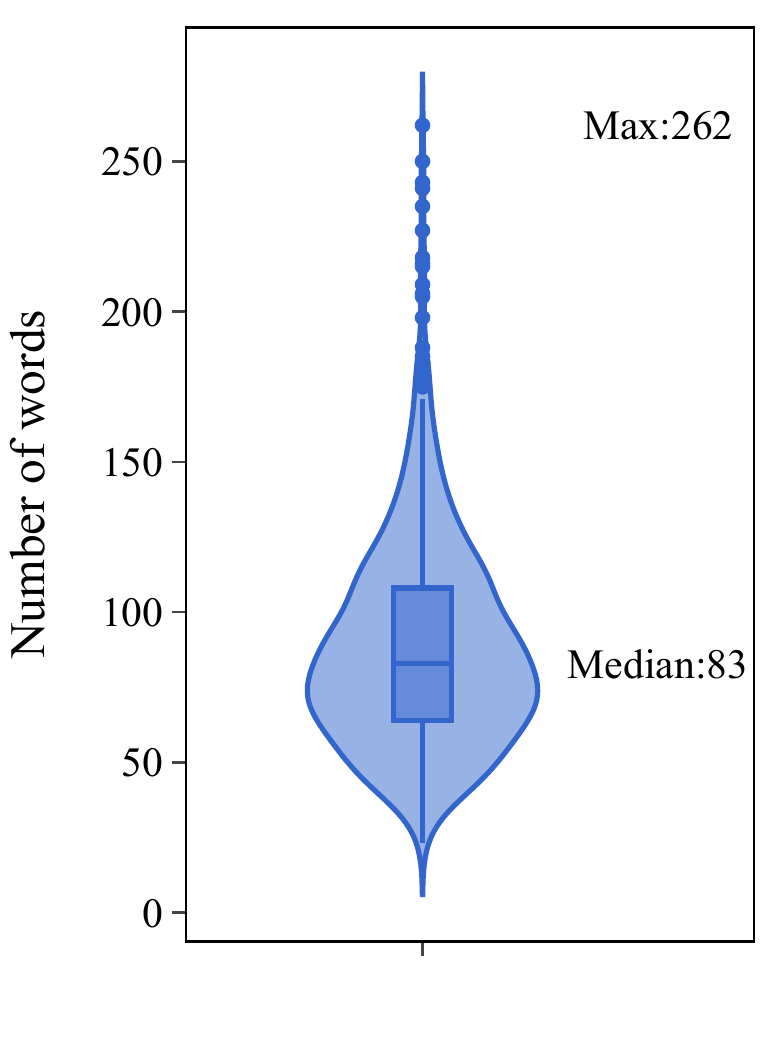}
    \caption{Question length.}
    \label{fig:question-length}
\end{subfigure}
\quad
\begin{subfigure}[t]{.22\linewidth}
  \centering
  \includegraphics[width=\linewidth]{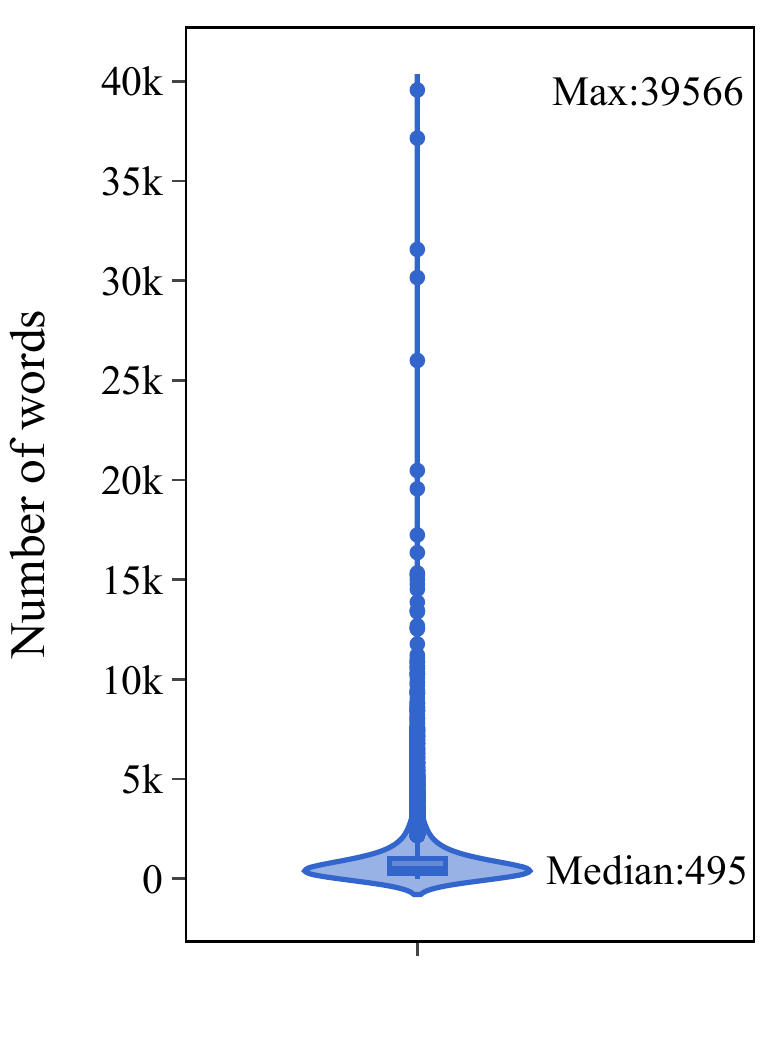}
  \caption{Article length.}
  \label{fig:article-length}
\end{subfigure}
\quad
\begin{subfigure}[t]{.22\linewidth}
    \centering
    \includegraphics[width=\linewidth]{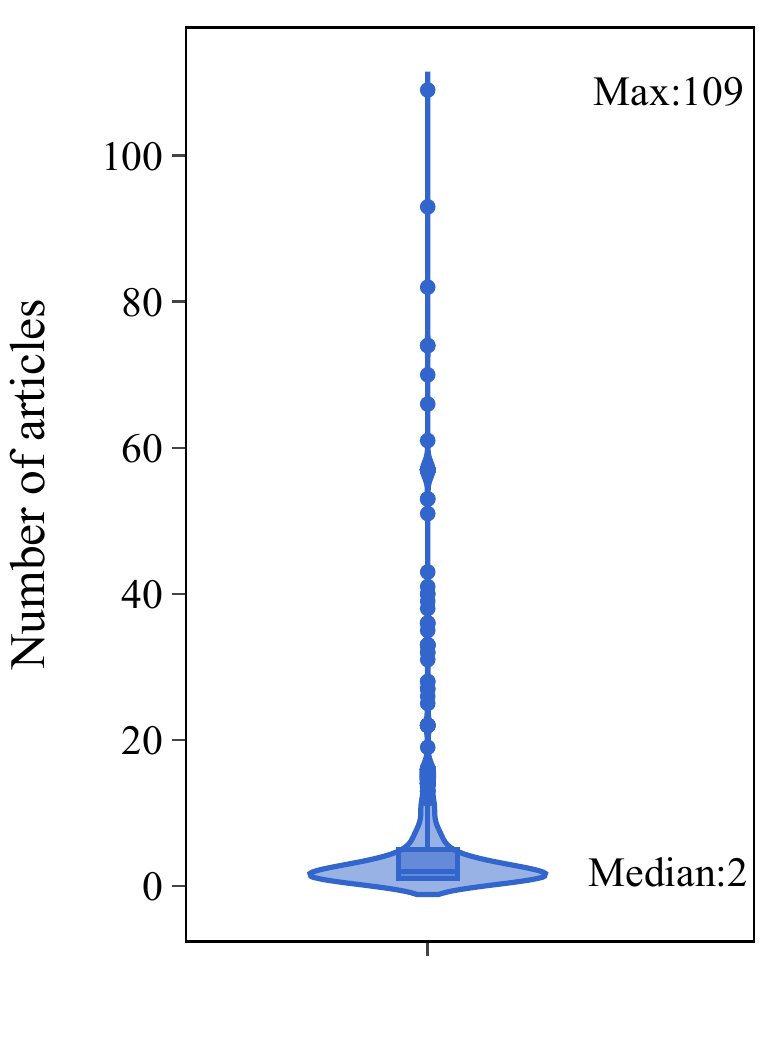}
    \caption{Number of relevant articles per question.}
    \label{fig:articles-per-question}
\end{subfigure}
\quad
\begin{subfigure}[t]{.22\linewidth}
    \centering
    \includegraphics[width=\linewidth]{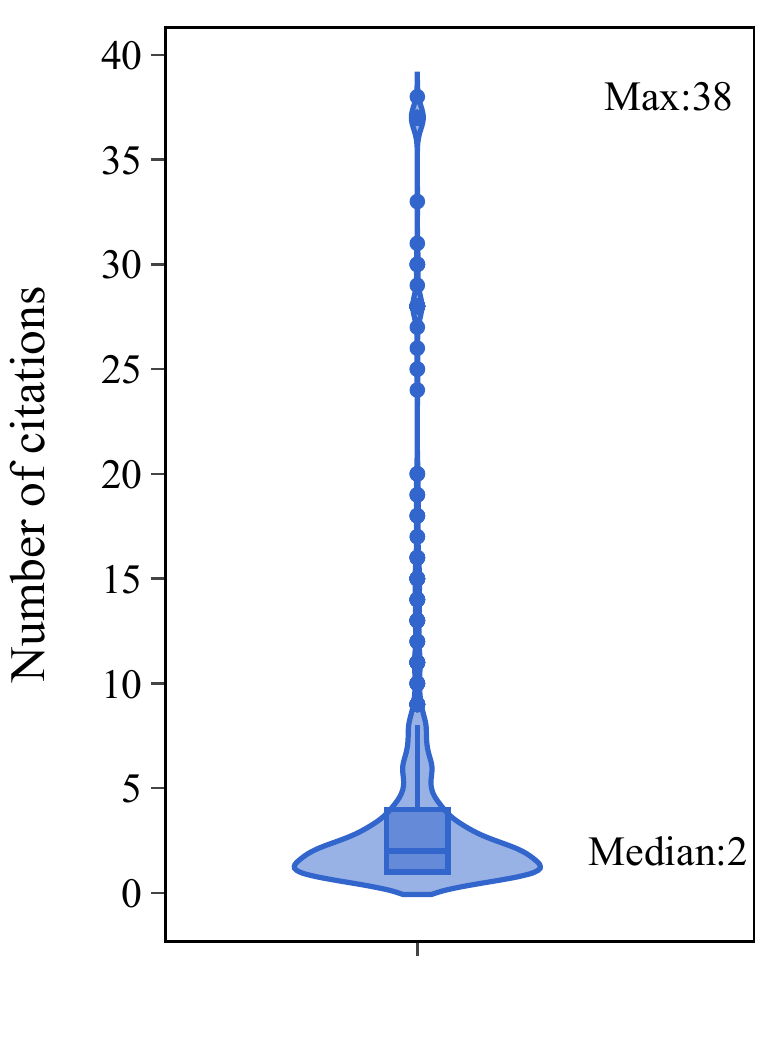}
    \caption{Number of citations per relevant article.}
    \label{fig:citations-per-article}
\end{subfigure}%
}
\caption{Statistics of BSARD.}
\label{fig:statistics-bsard}
\end{figure*}

\paragraph{Diversity.}
The 22,633 law articles that constitute our corpus have been collected from 32 Belgian codes covering a large number of legal topics, as presented in \Cref{tab:codes} of \Cref{app:legal_codes}. The articles have a median length of 495 words, but 25\% of them contain more than 1,026 words, and 40 articles exceed 10,000 words (the lengthiest one being up to 39,566 words), as illustrated in \Cref{fig:article-length}. These long articles are mostly \textit{general provisions}, i.e., articles that appear at the beginning of a code and define many terms and concepts later mentioned in the code. The questions are between 23 and 262 words long, with a median of 83 words, as shown in \Cref{fig:question-length}. They cover a wide range of topics, with around 85\% of them being either about family, housing, money, or justice, while the remaining 15\% concern either social security, foreigners, or work, as described in \Cref{tab:question-topics}.

\paragraph{Question-article relationship.}
Questions might have one or several relevant legal articles. Overall, 75\% of the questions have less than five relevant articles, 18\% have between 5 and 20, and the remaining 7\% have more than 20 with a maximum of 109, as seen in \Cref{fig:articles-per-question}. The latter often have complex and indirect answers that demand extensive reasoning over a whole code section, which explains these large numbers of relevant articles. Furthermore, an article deemed relevant to one question might also be for others. Therefore, we calculate for each unique article deemed relevant to at least one question the total number of times it is cited as a legal reference across all questions. As a result, we find that the median number of citations for those articles is 2, and less than 25\% of them are cited more than five times, as illustrated in \Cref{fig:citations-per-article}. Hence, out of the 22633 articles, only 1612 are referred to as relevant to at least one question in the dataset, and around 80\% of these 1612 articles come from either the Civil Code, Judicial Code, Criminal Investigation Code, or Penal Code. Meanwhile, 18 out of the 32 codes have less than five articles mentioned as relevant to at least one question, which can be explained by the fact that those codes focus less on individuals and their concerns.

\section{Models \label{sec:models}}
Formally speaking, a statutory article retrieval system $R: (q, \mathcal{C}) \rightarrow \mathcal{F}$ is a function that takes as input a question $q$ along with a corpus of law articles $\mathcal{C}$, and returns a much smaller filter set $\mathcal{F} \subset \mathcal{C}$ of the supposedly relevant articles, ranked by decreasing order of relevance. For a fixed $k=\left|\mathcal{F}\right|\ll|\mathcal{C}|$, the retriever can be evaluated in isolation with multiple rank-based metrics (see \Cref{subsec:experimental_setup}). The following section describes the retrieval models we use as a benchmark for the task.

\subsection{Lexical Models}
Traditionally, lexical approaches have been the de facto standard for textual information retrieval due to their robustness and efficiency. Given a query $q$ and an article $a$, a lexical model assigns to the pair $(q, a)$ a score $s_L: (q, a) \rightarrow \mathbb{R}_+$ by computing the sum, over the query terms, of the weights of each query term $t \in q$ in the article, i.e.,
\begin{equation}
    s_L(q, a) = \sum_{t \in q} w(t, a).
\end{equation}
First, we use the TF-IDF weighting scheme, in which
\begin{equation}
    w(t, a) = \operatorname{tf}(t, a) \cdot \log \frac{|\mathcal{C}|}{\operatorname{df}(t)},
\end{equation}
where the term frequency $\operatorname{tf}$ is the number of occurrences of term $t$ in article $a$, and the document frequency $\operatorname{df}$ is the number of articles within the corpus that contain term $t$. Then, we experiment with the BM25 weighting formula \citep{robertson1994okapi}, defined as
\begin{equation}
\begin{split}
    w(t, a)=&\frac{\operatorname{tf}(t, a){\cdot} (k_1{+}1)}{\operatorname{tf}(t, a){+}k_1{\cdot}\left(1{-}b{+}b\cdot \frac{|a|}{avgal}\right)}\\
            &\hspace{0.5cm}{\cdot}\log \frac{|\mathcal{C}|-\operatorname{df}(t)+0.5}{\operatorname{df}(t)+0.5},
\end{split}
\end{equation}
where $k_1 \in \mathbb{R}_+$ and $b \in [0,1]$ are constant parameters to be fixed, $|a|$ is the article length, and $avgal$ is the average article length in the collection.

During inference, we compute a score for each article in corpus $\mathcal{C}$ and return the $k$ articles with the highest scores as the top-$k$ most relevant results to the input query.

\subsection{Dense Models}
Lexical approaches suffer from the lexical gap problem \citep{berger2000bridging} and can only retrieve articles containing keywords present in the query. To overcome this limitation, recent work \citep{lee2019latent, karpukhin2020dense,xiong2021approximate} relies on neural-based architectures to capture semantic relationships between the query and documents. The most commonly used approach is based on a bi-encoder model \citep{gillick2018end} that maps queries and documents into dense vector representations. Formally, a dense retriever calculates a relevance score $s_D: (q,a) \rightarrow \mathbb{R}_+$ between question $q$ and article $a$ by the similarity of their respective embeddings $\boldsymbol{h}_q, \boldsymbol{h}_a \in \mathbb{R}^d$, i.e.,
\begin{equation}
    s_D(q,a) = \operatorname{sim}\left(\boldsymbol{h}_q, \boldsymbol{h}_a\right),
\end{equation}
where $\operatorname{sim}: \mathbb{R}^d \times \mathbb{R}^d \rightarrow \mathbb{R}$ is a similarity function such as dot product or cosine similarity. Typically, these embeddings result from a pooling operation on the output representations of a word embedding model:
\begin{equation}
\begin{split}
    \boldsymbol{h}_{q} &= \operatorname{pool}\left(f(q;\boldsymbol{\theta}_1)\right), \text{and}\\
    \boldsymbol{h}_a &= \operatorname{pool}\left(f(a;\boldsymbol{\theta}_2)\right),
\end{split}
\end{equation}
where model $f(\cdot; \boldsymbol{\theta}_i): \mathcal{W}^{n} \rightarrow \mathbb{R}^{n \times d}$ with parameters $\boldsymbol{\theta}_i$ maps an input text sequence of $n$ terms from vocabulary $\mathcal{W}$ to $d$-dimensional real-valued word vectors. The pooling operation $\operatorname{pool}: \mathbb{R}^{n \times d} \rightarrow \mathbb{R}^{d}$ uses the output word embeddings to distill a global representation for the text passage -- using either mean, max, or \texttt{[CLS]} pooling. 

Note that the bi-encoder architecture comes with two flavors: (i) \textsl{siamese} \citep{reimers2019sentence, xiong2021approximate}, which uses a unique word embedding model (i.e., $\boldsymbol{\theta}_1 = \boldsymbol{\theta}_2$) that maps the query and article together in a shared dense vector space, and (ii) \textsl{two-tower} \citep{yang2020multilingual, karpukhin2020dense}, which use two independent word embedding models that encode the query and article separately into different embedding spaces.

During inference, the articles are pre-encoded offline, and their representations are stored in an index structure. Then, given an input query, an exact search is performed by computing the similarities between the query representation and all pre-encoded article representations. The resulting scores are used to rank the articles such that the $k$ articles that have the highest similarities with the query are returned as the top-$k$ results.

\subsubsection{Zero-Shot Evaluation}
First, we study the effectiveness of siamese bi-encoders in a zero-shot evaluation setup, i.e., pre-trained word embedding models are applied out-of-the-box without any additional fine-tuning. We experiment with two types of widely-used word embedding models: (i) models that learned context-\textit{independent} word representations, namely word2vec \citep{mikolov2013efficient,mikolov2013distributed} and fastText \citep{bojanowski2017enriching}, and (ii) models that learned context-\textit{dependent} word embeddings, namely RoBERTa \citep{liu2019roberta}. 

RoBERTa can process texts up to a maximum input length of 512 tokens. Although alternative models exist to alleviate this limitation \citep{beltagy2020longformer, ainslie2020etc}, they have all been trained on English text, and there are no French equivalents available yet. Therefore, we use a simple workaround that splits the text into overlapping chunks and passes each chunk in turn to the embedding model. To form the chunks, we consider contiguous text sequences of 200 tokens with an overlap of 20 tokens between consecutive chunks.

For all zero-shot models, we use mean pooling on all word embeddings of the passage to extract a global representation for the latter and cosine similarity to score passage representations.

\subsubsection{Training}
Thereafter, we train our own siamese and two-tower RoBERTa-based bi-encoder models on BSARD. Let $\mathcal{D}=\{\langle q_{i}, a_{i}^{+}\rangle\}_{i=1}^{N}$ be the training data where each of the $N$ instances consists of a query $q_{i}$ associated with a relevant (positive) article $a_{i}^{+}$. Using in-batch negatives \citep{chen2017sampling, henderson2017efficient}, we can create a training set $\mathcal{T}=\{\langle q_{i}, a_{i}^{+}, \mathcal{A}_{i}^{-}\rangle\}_{i=1}^{N}$ where $\mathcal{A}_{i}^{-}$ is a set of negative articles for question $q_i$ constructed by considering the articles paired with the other questions from the same mini-batch. For each training instance, we contrastively optimize the negative log-likelihood of each positive article against their negative articles, i.e.,
\begin{equation}
\begin{aligned}
&L\left(q_{i}, a_{i}^{+}, \mathcal{A}_{i}^{-}\right) \\
=&-\log \frac{\operatorname{exp}\left(s_D(q_i,a_{i}^{+})/\tau\right)}
{\sum_{a \in \mathcal{A}_{i}^{-} \cup \{a_{i}^{+}\}} \operatorname{exp}\left(s_D(q_{i}, a)/\tau\right)},
\end{aligned}
\end{equation}
where $\tau > 0$ is a temperature parameter to be set. This contrastive loss allows learning embedding functions such that relevant question-article pairs will have a higher score than irrelevant ones.

To deal with articles longer than 512 tokens, we use the same workaround as in the zero-shot evaluation and split the long sequences into overlapping chunks of 200 tokens with a window size of 20. However, this time, we limit the size of the articles to the first 1000 words due to limited GPU memory. Although not ideal, doing so remains reasonable given that 75\% of the articles in our corpus have less than 1026 words, as mentioned in \Cref{subsec:dataset-analysis}. Each chunk is prefixed by the \texttt{[CLS]} token, and we extract a global representation for the whole article by averaging the output \texttt{[CLS]} token embeddings of the different chunks. Here, we use the dot product to compute similarities as it gives slightly better results than cosine.

\section{Experiments \label{sec:experiments}}
We now describe the setup we use for experiments and evaluate the performance of our models.

\subsection{Experimental Setup \label{subsec:experimental_setup}}
\paragraph{Metrics.}
We use three standard information retrieval metrics \citep{schutze2008introduction} to evaluate performance, namely the (macro-averaged) recall\at$k$ (R\at$k$), mean average precision\at$k$ (MAP\at$k$), and mean reciprocal rank\at$k$ (MRR\at$k$). \Cref{app:evaluation_metrics} gives a detailed description of these metrics in the context of statutory article retrieval. We deliberately omit to report the precision\at$k$ given that questions have a variable number of relevant articles (see \Cref{fig:articles-per-question}), which makes it senseless to report it at a fixed $k$ -- questions with $r$ relevant articles will always have P\at$k < 1$ if $k > r$. For the same reason, $k$ should be large enough for the recall\at$k$. Hence, we use $k \in \{100, 200, 500\}$ for our evaluation.

\paragraph{French word embedding models.}
Our focus is on a non-English dataset, so we experiment with French variants of the models mentioned above. Specifically, we use a 500-dimensional skip-gram word2vec model pre-trained on a crawled French corpus \citep{fauconnier2015french}, a 300-dimensional CBOW fastText model pre-trained on French Web data \citep{grave2018learning}, and a French RoBERTa model, namely CamemBERT \citep{martin2019camembert}, pre-trained on 147GB of French web pages filtered from Common Crawl.\footnote{\url{https://commoncrawl.org/}}

\paragraph{Hyper-parameters \& schedule.}
For BM25, we optimize the parameters on BSARD training set and find $k_1=1.0$ and $b=0.6$ to perform best. Regarding the bi-encoder models, we optimize the contrastive loss using a batch size of 22 question-article pairs and a temperature of 0.05 for 100 epochs, which is approximately 20,500 steps. We use AdamW \citep{loshchilov2019decoupled} with an initial learning rate of 2e-5, $\beta_1=0.9$, $\beta_2=0.999$, weight decay of 0.01, learning rate warm up over the first 500 steps, and linear decay of the learning rate. Training is performed on a single Tesla V100 GPU with 32 GBs of memory and evaluation on a server with a dual 20 core Intel(R) Xeon(R) E5-2698 v4 CPU \at 2.20GHz and 512 GBs of RAM.

\subsection{Results \label{subsec:results}}

\begin{table*}[t]
\centering
\resizebox{\textwidth}{!}{%
\begin{tabular}{cllcrrrrrr}
\hline
\textbf{Train} & \textbf{Model}       & \textbf{Encoder(s)} & \textbf{Params} & \textbf{Latency} & \textbf{R\at100} & \textbf{R\at200} & \textbf{R\at500} & \textbf{MAP\at100} & \textbf{MRR\at100} \\ \hline
\xmark         & TF-IDF               & -                   & -               & 827              & 40.13            & 50.44            & 59.34            & 8.69               & 12.98              \\
\xmark         & BM25 (official)      & -                   & -               & 1342             & 51.33            & 56.78            & 64.71            & 16.04              & 24.59              \\ \hline
\xmark         & Siamese bi-encoder   & word2vec            & -               & 4                & 49.41            & 61.76            & 71.57            & 12.90              & 21.49              \\
\xmark         & Siamese bi-encoder   & fastText            & -               & 3                & 32.93            & 41.33            & 49.26            & 6.29               & 11.78              \\
\xmark         & Siamese bi-encoder   & CamemBERT           & -               & 27               & 4.21             & 6.00             & 12.82            & 0.50               & 2.04               \\ \hline
\cmark         & Siamese bi-encoder   & CamemBERT           & 110M            & 28               & 71.63            & \textbf{78.38}   & \textbf{83.77}   & 35.44              & \textbf{43.52}     \\
\cmark         & Two-tower bi-encoder & CamemBERT           & 220M            & 26               & \textbf{74.78}   & 78.04            & 83.39            & \textbf{35.67}     & 42.46              \\ \hline
\end{tabular}%
}
\caption{Retrieval performance (in percent) and query latency (in milliseconds) of various information retrieval approaches on the test set. The best results are marked in bold.}
\label{tab:results}
\end{table*}

In \Cref{tab:results}, we report the retrieval performance of our models on the BSARD test set. Overall, the trained bi-encoder models significantly outperform all the other baselines. The two-tower model improves over its siamese variant on recall\at100 but performs similarly on the other metrics. Although BM25 underperforms the trained bi-encoders significantly, its performance indicates that it is still a strong baseline for domain-specific retrieval. These results are consistent with those obtained on other in-domain datasets \citep{thakur2021beir}.

Regarding the zero-shot evaluation of siamese bi-encoder models, we find that directly using the embeddings of a pre-trained CamemBERT model without optimizing for the IR task gives poor results. \citet{reimers2019sentence} noted similar findings for the task of semantic textual similarity. Furthermore, we observe that the word2vec-based bi-encoder significantly outperforms the fastText and BERT-based models, suggesting that pre-trained word-level embeddings are more appropriate for the task than character-level or subword-level embeddings when used out of the box.

Although promising, these results suggest ample opportunity for improvement compared to a skilled legal expert who can eventually retrieve all relevant articles to any question and thus get perfect scores.

\section{Discussion \label{sec:discussion}}
This section discusses the limitations and broader impacts of our dataset.

\subsection{Limitations}
As our dataset aims to give researchers a well-defined benchmark to evaluate existing and future legal information retrieval models, certain limitations need to be borne in mind to avoid drawing erroneous conclusions. 

First, the corpus of articles is limited to those collected from the 32 Belgian codes described in \Cref{tab:codes} of \Cref{app:legal_codes}, which does not cover the entire Belgian law as thousands of articles from decrees, directives, and ordinances are missing. During the dataset construction, all references to these uncollected articles are ignored, which causes some questions to end up with only a fraction of their initial number of relevant articles. This information loss implies that the answer contained in the remaining relevant articles might be incomplete, although it is still appropriate.

Additionally, it is essential to note that not all legal questions can be answered with statutes alone. For instance, the question “\textsl{Can I evict my tenants if they make too much noise?}” might not have a detailed answer within the statutory law that quantifies a specific noise threshold at which eviction is allowed. Instead, the landlord should probably rely more on case law and find precedents similar to their current situation (e.g., the tenant makes two parties a week until 2 am). Hence, some questions are better suited than others to the statutory article retrieval task, and the domain of the less suitable ones remains to be determined.

\subsection{Broader Impacts}
In addition to helping advance the state-of-the-art in retrieving statutes relevant to a legal question, BSARD-based models could improve the efficiency of the legal information retrieval process in the context of legal research, therefore enabling researchers to devote themselves to more thoughtful parts of their research.

Furthermore, BSARD can become a starting point of new \textit{open-source} legal information search tools so that the socially weaker parties to disputes can benefit from a free professional assisting service. However, there are risks that the dataset will not be used exclusively for the public interest but perhaps also for profit as part of proprietary search tools developed by companies. Since this would reinforce rather than solve the problem of access to legal information and justice for all, we decided to distribute BSARD under a license with a non-commercial clause.

Other potential negative societal impacts could involve using models trained on BSARD to misuse or find gaps within the governmental laws or use the latter not to defend oneself but to deliberately damage people or companies instead. Of course, we discourage anyone from developing models that aim to perform the latter actions.

\section{Conclusion \label{sec:conclusion}}
In this paper, we present the Belgian Statutory Article Retrieval Dataset (BSARD), a citizen-centric French native dataset for statutory article retrieval. Within a larger effort to bridge the gap between people and the law, BSARD provides a means of evaluating and developing models capable of retrieving law articles relevant to a legal question posed by a layperson. We benchmark several strong information retrieval baselines that show promise for the feasibility of the task yet indicate room for improvement. In the future, we plan to build retrieval models that can handle lengthy statutory articles and inherently exploit the hierarchy of the law. In closing, we hope that our work sparks interest in developing practical and reliable statutory article retrieval models to help improve access to justice for all.


\section*{Acknowledgments}
This research is partially supported by the Sector Plan Digital Legal Studies of the Dutch Ministry of Education, Culture, and Science. In addition, this research was made possible, in part, using the Data Science Research Infrastructure (DSRI) hosted at Maastricht University.

\bibliographystyle{packages/acl_natbib}
\bibliography{refs.bib}

\appendix
\section*{Appendix}

\begin{table*}[ht!]
\centering
\begin{tabular}{llll}
\hline
Authority & Code                                                & \#Articles & \#Relevant \\ \hline
Federal   & Judicial Code                                       & 2285       & 429        \\
          & Code of Economic Law                                & 2032       & 98         \\
          & Civil Code                                          & 1961       & 568        \\
          & Code of Workplace Welfare                           & 1287       & 25         \\
          & Code of Companies and Associations                  & 1194       & 0          \\
          & Code of Local Democracy and Decentralization        & 1159       & 3          \\
          & Navigation Code                                     & 977        & 0          \\
          & Code of Criminal Instruction                        & 719        & 155        \\
          & Penal Code                                          & 689        & 154        \\
          & Social Penal Code                                   & 307        & 23         \\
          & Forestry Code                                       & 261        & 0          \\
          & Railway Code                                        & 260        & 0          \\
          & Electoral Code                                      & 218        & 0          \\
          & The Constitution                                    & 208        & 5          \\
          & Code of Various Rights and Taxes                    & 191        & 0          \\
          & Code of Private International Law                   & 135        & 4          \\
          & Consular Code                                       & 100        & 0          \\
          & Rural Code                                          & 87         & 12         \\
          & Military Penal Code                                 & 66         & 1          \\
          & Code of Belgian Nationality                         & 31         & 8          \\ \hline
Regional  & Walloon Code of Social Action and Health            & 3650       & 40         \\
          & Walloon Code of the Environment                     & 1270       & 22         \\
          & Walloon Code of Territorial Development             & 796        & 0          \\
          & Walloon Public Service Code                         & 597        & 0          \\
          & Walloon Code of Agriculture                         & 461        & 0          \\
          & Brussels Spatial Planning Code                      & 401        & 1          \\
          & Walloon Code of Basic and Secondary Education       & 310        & 0          \\
          & Walloon Code of Sustainable Housing                 & 286        & 20         \\
          & Brussels Housing Code                               & 279        & 44         \\
          & Brussels Code of Air, Climate and Energy Management & 208        & 0          \\
          & Walloon Animal Welfare Code                         & 108        & 0          \\
          & Brussels Municipal Electoral Code                   & 100        & 0          \\ \hline
Total     &                                                     & 22633      & 1612       \\
\hline
\end{tabular}
\caption{Summary of the number of articles collected (after pre-processing) from each of the Belgian codes considered for BSARD, as well as the number of articles found to be relevant for at least one of the legal questions.}
\label{tab:codes}
\end{table*}

\section{Legal Codes\label{app:legal_codes}}
\Cref{tab:codes} presents a detailed summary of the 32 publicly available Belgian codes collected for BSARD.

\section{Evaluation Metrics \label{app:evaluation_metrics}}
Let $\operatorname{rel}_{q}(a) \in \{0,1\}$ be the binary relevance label of article $a$ for question $q$, and $\langle i, a\rangle \in \mathcal{F}_q$ a result tuple (article $a$ at rank $i$) from the filter set $\mathcal{F}_q \subset \mathcal{C}$ of ranked articles retrieved for question $q$.

\paragraph{Recall.}
The \textit{recall} $\operatorname{R}_{q}$ is the fraction of relevant articles retrieved for query $q$ w.r.t. the total number of relevant articles in the corpus $\mathcal{C}$, i.e.,
\begin{equation}
    \operatorname{R}_{q}=\frac{\sum_{\langle i, a\rangle \in \mathcal{F}_q} \operatorname{rel}_{q}(a)}{\sum_{a \in \mathcal{C}} \operatorname{rel}_{q}(a)}.
\end{equation}

\paragraph{Reciprocal rank.}
The \textit{reciprocal rank} ($\operatorname{RR}_{q}$) calculates the reciprocal of the rank at which the first relevant article is retrieved, i.e.,
\begin{equation}
    \operatorname{RR}_{q}=\max _{\langle i, a\rangle \in \mathcal{F}_q} \frac{\operatorname{rel}_{q}(a)}{i}.
\end{equation}

\paragraph{Average precision.}
The \textit{average precision} $\operatorname{AP}_{q}$ is the mean of the precision value obtained after each relevant article is retrieved, that is
\begin{equation}
    \operatorname{AP}_{q}=\frac{\sum_{\langle i, a\rangle \in \mathcal{F}_q} \operatorname{P}_{q, i} \times \operatorname{rel}_{q}(a)}{\sum_{a \in \mathcal{C}} \operatorname{rel}_{q}(a)},
\end{equation}
where $\operatorname{P}_{q, j}$ is the \textit{precision} computed at rank $j$ for query $q$, i.e., the fraction of relevant articles retrieved for query $q$ w.r.t. the total number of articles in the retrieved set $\{\mathcal{F}_q\}_{i=1}^{j}$:
\begin{equation}
    \operatorname{P}_{q, j}=\frac{\sum_{\langle i, a\rangle \in \{\mathcal{F}_q\}_{i=1}^{j}} \operatorname{rel}_{q}(a)}{\left|\{\mathcal{F}_q\}_{i=1}^{j}\right|}.
\end{equation}

We report the macro-averaged \textit{recall} (R), \textit{mean reciprocal rank} (MRR), and \textit{mean average precision} (MAP), which are the average values of the corresponding metrics over a set of $n$ queries. Note that as those metrics are computed for a filter set of size $k=\left|\mathcal{F}_q\right|\ll|\mathcal{C}|$ (and not on the entire list of articles in $\mathcal{C}$), we report them with the suffix ``\at\textit{k}''.

\section{Dataset Documentation}

\begin{table*}[t]
\centering
\small
\begin{subtable}[t]{.55\linewidth}
\centering
\caption*{}
    \begin{tabular}{|lr|}
        \hline
        \multicolumn{2}{|l|}{\textbf{\Large Data Facts}}                                      \\
        \multicolumn{2}{|l|}{Belgian Statutory Article Retrieval Dataset (BSARD)}             \\ \hline
        \multicolumn{2}{l}{}                                                                  \\ \hline
        \multicolumn{2}{|l|}{\textbf{\large Metadata}}                                        \\ \hline
        \textbf{Filename}    & articles\_fr.csv$^\ast$                                        \\
                             & questions\_fr\_train.csv$^\dagger$                             \\
                             & questions\_fr\_test.csv$^\ddagger$                             \\ \hline
        \textbf{Format}      & CSV                                                            \\ \hline
        \textbf{Url}         & {\scriptsize\url{https://doi.org/10.5281/zenodo.5217310}}  \\ \hline
        \textbf{Domain}      & natural language processing                                    \\ \hline
        \textbf{Keywords}    & information retrieval, law                                     \\ \hline
        \textbf{Type}        & tabular                                                        \\ \hline
        \textbf{Rows}        & 22633$^\ast$, 886$^\dagger$, 222$^\ddagger$                    \\ \hline
        \textbf{Columns}     & 6$^\ast$, 6$^\dagger$, 6$^\ddagger$                            \\ \hline
        \textbf{Missing}     & none                                                           \\ \hline
        \textbf{License}     & CC BY-NC-SA 4.0                                                \\ \hline
        \textbf{Released}    & August 2021                                                    \\ \hline
        \textbf{Range}       & N/A.                                                           \\ \hline
        \textbf{Description} & This dataset is a collection of French                         \\
                             & native legal questions posed by Belgian                      \\
                             & citizens and law articles from the                              \\
                             & Belgian legislation. The articles come                         \\
                             & from 32 publicly available Belgian                        \\
                             & codes. Each question is labeled by one                          \\
                             & or several relevant articles from the                             \\
                             & corpus. The annotations were done by                            \\
                             & a team of experienced Belgian jurists.                         \\\hline
        \multicolumn{2}{l}{}                                                                                                        \\ \hline
        \multicolumn{2}{|l|}{\textbf{\large Provenance}}                                                                            \\ \hline
        \multicolumn{2}{|l|}{\textbf{Source}}                                                                                       \\
        \multicolumn{2}{|l|}{Belgian legislation}                                                                                   \\
        \multicolumn{2}{|l|}{\scriptsize\hspace{0.5cm}(\url{https://www.ejustice.just.fgov.be/loi/loi.htm})}                        \\
        \multicolumn{2}{|l|}{Droits Quotidiens}                                                                                     \\
        \multicolumn{2}{|l|}{\scriptsize\hspace{0.5cm}(\url{https://droitsquotidiens.be})}                                      \\ \hline
        \textbf{Author}      &                                                                                                      \\
        Name                 & Antoine Louis                                                                                 \\
        Email                & \href{mailto:}{\texttt{\scriptsize a.louis@maastrichtuniversity.nl}}  \\ \hline
    \end{tabular}
\end{subtable}%
\begin{subtable}[t]{.45\linewidth}
\centering
\caption*{}
    \begin{tabular}{|lr|}
        \hline
        \multicolumn{2}{|l|}{\textbf{\large Variables}}                                                                                  \\ \hline
        \textbf{id$^\ast$}                     & \begin{tabular}[c]{@{}r@{}}A unique ID number\\ for the article.\end{tabular}    \\ \hline
        \textbf{article$^\ast$}                & \begin{tabular}[c]{@{}r@{}}The full content\\ of the article.\end{tabular}       \\ \hline
        \textbf{code$^\ast$}                   & \begin{tabular}[c]{@{}r@{}}The code to which\\ the article belongs.\end{tabular} \\ \hline
        \textbf{article\_no$^\ast$}            & \begin{tabular}[c]{@{}r@{}}The article number\\ in the code.\end{tabular}        \\ \hline
        \textbf{description$^\ast$}            & \begin{tabular}[c]{@{}r@{}}The concatenated headings\\ of the article.\end{tabular}        \\ \hline
        \textbf{law\_type$^\ast$}              & \begin{tabular}[c]{@{}r@{}}Either "regional" or\\ "national" law.\end{tabular}   \\ \hline
        \textbf{id$^{\dagger,\ddagger}$}       & \begin{tabular}[c]{@{}r@{}}A unique ID number\\ for the question.\end{tabular}   \\ \hline
        \textbf{question$^{\dagger,\ddagger}$} & \begin{tabular}[c]{@{}r@{}}The content of\\ the question.\end{tabular}           \\ \hline
        \textbf{category$^{\dagger,\ddagger}$}     & \begin{tabular}[c]{@{}r@{}}The general topic\\ of the question.\end{tabular}              \\ \hline
        \textbf{subcategory$^{\dagger,\ddagger}$}  & \begin{tabular}[c]{@{}r@{}}The precise topic\\ of the question.\end{tabular}              \\ \hline
        \textbf{extra\_description$^{\dagger,\ddagger}$}  & \begin{tabular}[c]{@{}r@{}}Extra categorization\\ tags of the question.\end{tabular}          \\ \hline
        \textbf{article\_ids$^{\dagger,\ddagger}$} & \begin{tabular}[c]{@{}r@{}}A list of article IDs\\ relevant to the question.\end{tabular} \\ \hline
    \end{tabular}
\end{subtable}
\caption{Dataset nutrition labels for BSARD.}
\label{tab:data_nutrition_labels}
\end{table*}

\subsection{Dataset Nutrition Labels}
As a first way to document our dataset, we provide the \textit{dataset nutrition labels} \citep{holland2018dataset} for BSARD in \Cref{tab:data_nutrition_labels}. 

\subsection{Data Statement}
In addition to the data nutrition labels, we include the \textit{data statement} \citep{bender2018data} for BSARD, which provides detailed context on the dataset so that researchers, developers, and users can understand how models built upon it might generalize, be appropriately deployed, and potentially reflect bias or exclusion.

\paragraph{Curation rationale.} 
All law articles from the selected Belgian codes were included in our dataset, except those revoked (identifiable because mentioned before the article or empty content) and those with a duplicate number within the same code (namely, the articles from Act V, Book III of the Civil Code; from Sections 2, 2bis, and 3 of Chapter II, Act VIII, Book III of the Civil Code; from Act XVIII, Book III of the Civil Code; from the Preliminary Act of the Code of Criminal Instruction; from the Appendix of the Judicial Code). Not including the latter articles did not pose a vital concern because none of them were mentioned as relevant to any of the questions in our dataset. Regarding the questions, all those that referenced at least one of the articles from our corpus were included in the dataset.

\paragraph{Language variety.}
The questions and legal articles were collected in French (fr-BE) as spoken in Wallonia and Brussels-Capital region.

\paragraph{Speaker demographic.}
Speakers were not directly approached for inclusion in this dataset and thus could not be asked for demographic information. Questions were collected, anonymized, and reformulated by Droits Quotidiens. Therefore, no direct information about the speakers’ age and gender distribution or socioeconomic status is available. However, it is expected that most, but not all, of the speakers are adults (18+ years), speak French as a native language, and live in Wallonia or Brussels-Capital region.

\paragraph{Annotator demographic.}
A total of six Belgian jurists from Droits Quotidiens contributed to annotating the questions. All have a law degree from a Belgian university and years of experience in providing legal advice and clarifications of the law. They range in age from 30-60 years, including one man and five women, gave their ethnicity as white European, speak French as a native language, and represent upper middle class based on income levels.

\paragraph{Speech situation.}
All the questions were written between 2018 and 2021 and collected in May 2021. They represent informal, asynchronous, edited, written language that does not exceed 265 words. None of the questions contained hateful, aggressive, or inappropriate language as they were all reviewed and reworded by Droits Quotidiens to be neutral, anonymous, and comprehensive. All the legal articles were written between 1804 and 2021 and collected in May 2021. They represent strong, formal, written language containing up to 39,570 words.

\paragraph{Text characteristics.}
Many articles complement or rely on other articles in the same or another code and thus contain (sometimes lengthy) legal references, which might be seen as noisy data.

\paragraph{Recording quality.}
N/A.

\paragraph{Other.}
N/A.

\paragraph{Provenance appendix.}
N/A.

\subsection{Intended Uses}
The dataset is intended to be used by researchers to build and evaluate models on retrieving law articles relevant to an input legal question. Therefore, it should not be regarded as a reliable source of legal information at this point in time, as both the questions and articles correspond to an outdated version of the Belgian law from May 2021 (time of dataset collection). In the latter case, the user is advised to consult daily updated official legal resources (e.g., the Belgian Official Gazette).

\subsection{Hosting}
We provide access to BSARD on Hugging Face Datasets \citep{lhoest2021datasets} at {\footnotesize \url{https://huggingface.co/datasets/antoiloui/bsard}}. Additionally, the dataset is hosted on Zenodo at {\footnotesize \url{https://doi.org/10.5281/zenodo.5217310}}.

\subsection{Data Format}
The dataset is stored as CSV files and can be read using standard libraries (e.g., the built-in \texttt{csv} module in Python) or the \href{https://pypi.org/project/datasets/}{\hf \texttt{datasets}} library:
\begin{lstlisting}[style=CEE]
from datasets import load_dataset
data = load_dataset("antoiloui/bsard")
\end{lstlisting}

\subsection{Reproducibility}
We ensure the reproducibility of the experimental results by releasing our code on Github at {\footnotesize \url{https://github.com/maastrichtlawtech/bsard}}.

\subsection{Licensing}
The dataset is publicly distributed under a \href{https://creativecommons.org/licenses/by-nc-sa/4.0/}{CC BY-NC-SA 4.0} license, which allows sharing freely (i.e., copy and redistribute) and adapt (i.e., remix, transform, and build upon) the material on the conditions that the latter is used for non-commercial purposes only, proper attribution is given (i.e., appropriate credit, link to the license, and an indication of changes), and the same license as the original is used if one distributes an adapted version of the material. In addition, the code to reproduce the experimental results of the paper is released under the MIT license.

\subsection{Maintenance}
The dataset will be supported and maintained by the Law \& Tech Lab at Maastricht University. Any updates to the dataset will be communicated via the Github repository. All questions and comments about the dataset can be sent to Antoine Louis: \href{mailto:}{\footnotesize \texttt{a.louis@maastrichtuniversity.com}}. Other contacts can be found at {\footnotesize \url{https://maastrichtuniversity.nl/law-and-tech-people}}.

\end{document}